\documentclass[review]{elsarticle}
\usepackage{lineno,hyperref}
\journal{Journal of Medical Image Analysis}
\biboptions{authoryear}
\usepackage{caption}
\usepackage{color, soul}
\begin{document}
\begin{frontmatter}
\title{A Real-time and Registration-free Framework for Dynamic Shape Instantiation}
\author{Xiao-Yun Zhou}
\ead{xiaoyun.zhou14@imperial.ac.uk}
\author{Guang-Zhong Yang}
\ead{g.z.yang@imperial.ac.uk}
\author{Su-Lin Lee\corref{mycorrespondingauthor}}
\cortext[mycorrespondingauthor]{Corresponding author}
\ead{su-lin.lee@imperial.ac.uk}
\address{The Hamlyn Centre for Robotic Surgery, Imperial College London, London, UK}

\begin{abstract}
Real-time 3D navigation during minimally invasive procedures is an essential yet challenging task, especially when considerable tissue motion is involved. To balance image acquisition speed and resolution, only 2D images or low-resolution 3D volumes can be used clinically. In this paper, a real-time and registration-free framework for dynamic shape instantiation, generalizable to multiple anatomical applications, is proposed to instantiate high-resolution 3D shapes of an organ from a single 2D image intra-operatively. Firstly, an approximate optimal scan plane was determined by analyzing the pre-operative 3D statistical shape model (SSM) of the anatomy with sparse principal component analysis (SPCA) and considering practical constraints . Secondly, kernel partial least squares regression (KPLSR) was used to learn the relationship between the pre-operative 3D SSM and a synchronized 2D SSM constructed from 2D images obtained at the approximate optimal scan plane. Finally, the derived relationship was applied to the new intra-operative 2D image obtained at the same scan plane to predict the high-resolution 3D shape intra-operatively. A major feature of the proposed framework is that no extra registration between the pre-operative 3D SSM and the synchronized 2D SSM is required. Detailed validation was performed on studies including the liver and right ventricle (RV) of the heart. The derived results (mean accuracy of $2.19mm$ on patients and computation speed of $1ms$) demonstrate its potential clinical value for real-time, high-resolution, dynamic and 3D interventional guidance. 
\end{abstract}

\begin{keyword}
Image-guided interventions\sep Dynamic shape instantiation\sep 3D shape recovery\sep KPLSR
\end{keyword}
\end{frontmatter}
%\linenumbers

\section{Introduction}
Current clinical systems for minimally invasive procedures, such as cardiac radio-frequency ablation, image-guided needle biopsies, and endovascular interventions, typically incorporate static 3D surfaces for guidance. Real-time dynamic tracking of 3D surfaces can help to optimize the interventional procedure, especially for complex anatomical structures undergoing gross tissue deformation, bulk organ motion, and potential topological changes during interventions.

A combination of multiple imaging modalities has been used for dynamic 3D navigation. For example, a real-time registration scheme based on both spatial registration and electrocardiography was proposed to overlay pre-operative 3D magnetic resonance (MR) or computed tomography (CT) volumes onto intra-operative 2D ultrasound images for dynamic 3D navigation \citep{huang2009dynamic}. 3D transesophageal echocardiography (TEE) was fused with 2D X-ray fluoroscopic images using image localization and calibration for dynamic cardiac navigation \citep{gao2012registration}. However, based on a combination of multiple imaging modalities, the dynamic 3D shapes were either interpolated from pre-operative 3D volumes or intra-operatively collected 3D volumes with low-resolution. A 3D shape recovery scheme based on intra-operative 2D images including X-ray, ultrasound, and MR could take intra-operative information into account whilst achieving high-resolution at the same time. This kind of 3D shape recovery is termed dynamic shape instantiation. The scheme may or may not involve the use of template models \citep{filippi2008analysis}. Without template models used, more intra-operative information and longer image acquisition time are needed; for example, at least seven intra-operative 2D images were needed for reasonable 3D prostate reconstruction \citep{cool20063d}. In this paper, a single intra-operative 2D image is targeted and hence we focus on template-based 3D shape instantiation.

For template-based 3D shape instantiation methods, statistical shape models (SSM) \citep{frangi2002automatic}, free form deformation (FFD) \citep{koh2011reconstruction}, and Laplacian surface deformation \citep{karade20153d} can be used for the representation of templates. SSM \citep{cootes1995active} is a popular technique which represents a set of 3D meshes or 2D contours with the same number of vertices and connectivities. SSM-based 3D shape instantiation learns from shape variations rather than only applying smoothness and 2D/3D similarity as the constraints. It deforms an initial 3D SSM to match intra-operative sparse inputs such as ultrasound-derived surface points \citep{barratt2008instantiation}, digitized landmarks \citep{rajamani2007statistical}, or two or more calibrated X-ray images \citep{baka20112d}. These methods usually learn a model from a training set of anatomies of multiple patients and deform the learned model for a new patient, which requires a high anatomical similarity between patients. This learning is not suitable for patients with anatomical anomalies. For example, patients who have undergone liver resection have a significantly different liver shape to other subjects. A possible solution for these specific cases has been proposed  in \citep{lee2010dynamic}. Here, limited optimal scan planes were determined by analyzing the pre-operative and patient-specific 3D SSM of the liver with principal component analysis (PCA). The relationship between pre-operative 3D SSM and synchronized 2D SSM constructed from 2D images at the optimal scan planes was learned by partial least squares regression (PLSR). Finally, with new intra-operative 2D images obtained at the same scan planes, the 3D shape was instantiated intra-operatively by applying the PLSR-derived relationship. However, in \citep{lee2010dynamic}, the optimal scan plane determination depended on the selected vertices that were deemed informative but were highly correlated and clustered. PLSR can only derive linear relationships while the deformations of most anatomies are non-linear. Based on \citep{lee2010dynamic}, a framework which achieves more accurate, robust, generalizable and convenient shape instantiations from a single intra-operative 2D image is proposed in this paper.

Subspace reprojection was proposed to determine an optimal scan plane for SSM-based 3D shape instantiation by fitting a plane to the most informative vertices \citep{lee2005assessment}. This optimal scan plane was shown to have enhanced accuracy compared to other scan planes\citep{lee2005assessment}. By applying PCA \citep{jolliffe2002principal} on the pre-operative 3D SSM, the informative vertices which contribute most to the shape variations are determined by the loadings of principal components \citep{lee2010dynamic}. The downside of using PCA is that the derived principal components are linear combinations of multiple variables and therefore the selected informative variables are highly related and difficult to interpret. This phenomenon when reflected in our application is that the selected informative vertices are clustered and are not the real and independent informative vertices. Many methods have been proposed to solve this issue, including rotation methods \citep{jolliffe1995rotation}, limited set of integers \citep{vines2000simple}, and simplified component technique least absolute shrinkage and selection operator (SCoTLASS) \citep{jolliffe2003modified}. Simple thresholding is a common and informal method usually used in practice \citep{lee2010dynamic}; however, this method lacks theoretical support and usually causes problems \citep{cadima1995loading}. Recently, Zou et al. proposed sparse PCA (SPCA) which reformulated PCA into a regression-type optimization problem and then added a L1 constraint to achieve sparse loadings; they demonstrated improved performance of SPCA in selecting the real informative variables over previous methods \citep{zou2006sparse}. A SPCA toolbox was later developed \citep{sjostrand2012spasm}.

PLSR is a linear regression method which has a similar prediction accuracy to ridge regression (RR) and principal component regression (PCR) \citep{frank1993statistical}. It is more widely used than RR and PCR in medical problems, such as cardiac motion prediction \citep{ablitt2004predictive} and craniofacial reconstruction \citep{duan20153d}, as it is more suitable for problems with a larger number of variables and fewer number of observations \citep{rosipal2001kernel}. However, its accuracy for non-linear motions is limited.

Many non-linear PLSR variations have been proposed and they can be divided into two groups \citep{rosipal2006overview}: the first group reformulates the linear relationship into a non-linear one by polynomial functions, smoothing splines, artificial neural networks, and radial basis function networks while the second group maps the original variables into a higher dimensional space and regresses the mapped variables in the higher dimension, for example, kernel space. Kernel PLSR (KPLSR)  \citep{rosipal2001kernel} from the second group is adopted in this paper for improved computation speed as its formulation is as time-efficient as PLSR and avoids the non-linear optimization in the first group.

\begin{figure}[thpb]
\centering
\includegraphics[width=1.0\textwidth]{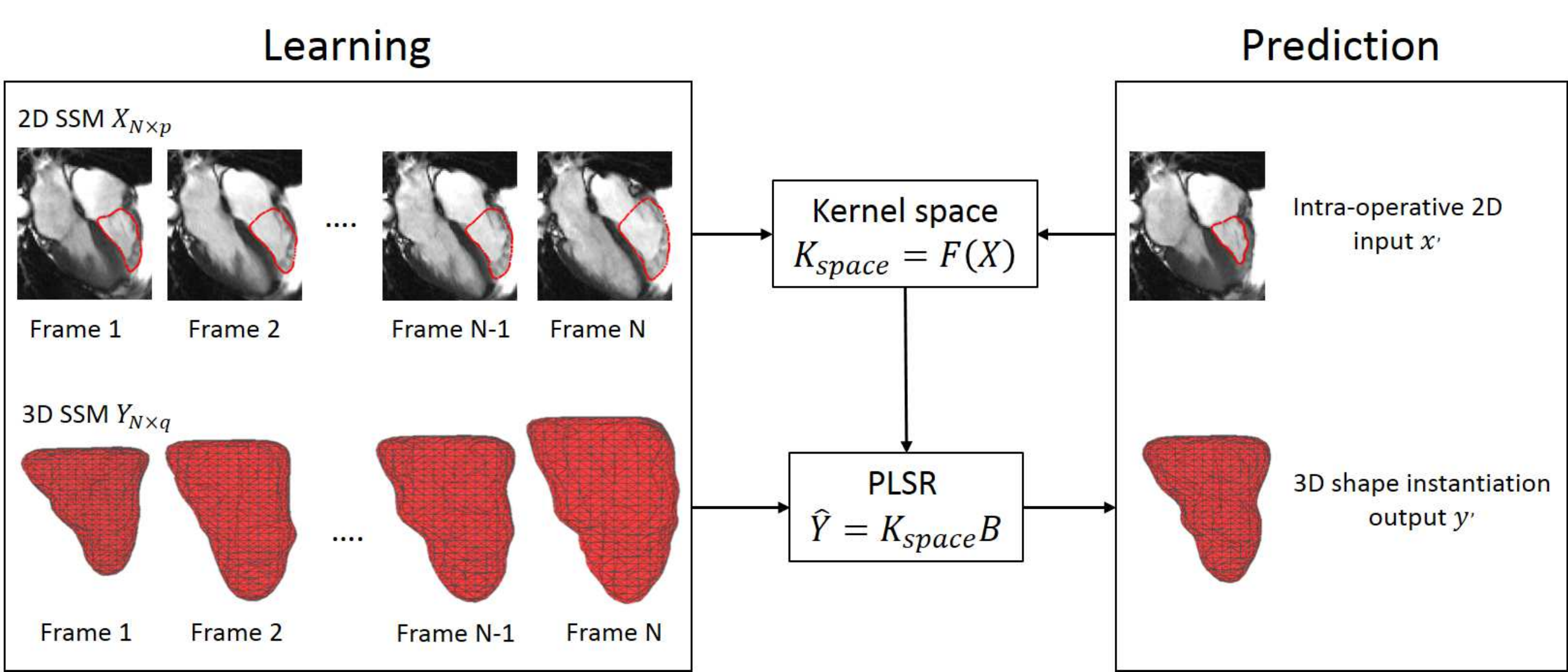}
\caption{A schematic illustration of the overall framework of the proposed dynamic shape instantiation scheme: both the 2D images in the learning and prediction are taken at the approximate optimal scan plane; the learning 2D SSM and learning 3D SSM are not registered but synchronized.}
\label{fig:frame_work}
\end{figure}

In this paper, the high-resolution 3D shape of a dynamic anatomy was instantiated from a single intra-operative 2D image in real-time. Firstly, the anatomy was scanned by MR or CT pre-operatively for multiple 3D volumes along the dynamic cycle and a 3D SSM was constructed. SPCA was applied on the pre-operative 3D SSM to select the informative vertices which were used to fit an optical scan plane. Local adjustments of the scan plane parameters for better accessibility, visibility or satisfying other local constraints is possible without incurring major errors, as the later KPLSR-based 3D shape instantiation scheme is robust to optimal scan plane derivations. Secondly, 2D images synchronized with the pre-operative scanning were obtained at the approximate optimal scan plane and were sampled to generate a synchronized 2D SSM. KPLSR was applied to learn the relationship between the pre-operative 3D SSM and the synchronized 2D SSM. Finally, the high-resolution 3D shape was instantiated intra-operatively by applying the KPLSR-derived relationship onto a new intra-operative 2D image at the same scan plane. The overall framework of the proposed dynamic shape instantiation is illustrated in Fig. \ref{fig:frame_work}. Due to the learning of patient-specific models, the framework is applicable to any anatomy. No extra registration is needed for the pre-operative 3D SSM and the synchronized 2D SSM. Validation was performed on the liver (two digital liver phantoms, one dynamic liver phantom, one in vivo porcine liver, eight metastatic livers) and the cardiac right ventricle (RV) (18 asymptomatic RVs and 9 hypertrophic cardiomyopathy (HCM) RVs); we anticipate that potential applications of our work will include percutaneous liver biopsy, cardiac catheterization \citep{razavi2003cardiac}, and intra-myocardial therapy \citep{saeed2005mri}. For example, in cardiac ablation, the instantiated 3D RV shape can be used to help navigate the catheter tip to the target ablation area.

\section{Methodology}
The methods for determining the optimal scan plane are described in Sec. \ref{Sec: Optimal Scan Plane}. The learning and instantiation based on KPLSR are described in Sec. \ref{Sec: Prediction}. Finally, the data collection and detailed validation experiments are in Sec. \ref{Sec: Data}.

\subsection{Optimal Scan Plane Determination}
\label{Sec: Optimal Scan Plane}
By pre-operatively scanning the target anatomy with CT or MR, a 4D volume consisting of multiple 3D volumes at different time frames along the dynamic cycle of the anatomy was obtained. These 3D volumes were represented with 3D meshes using the same number of vertices and connectivities, which created a pre-operative 3D SSM (a point distribution model) with vertices $Y_{N\times numY\times 3}$, where $N$ is the number of time frames and $numY$ is the number of vertices. By rearranging the $(x, y, z)$ coordinates of the vertices as independent variables, $Y_{N\times q}$ was obtained, where $q=numY\times 3$ is the number of variables. Without loss of generality, $Y_{N\times q}$ was centered and normalized as $Y_{norm}$ with the mean and norm of each column as 0 and 1.

For data $Y_{norm}$, its singular value decomposition is $Y_{norm}=UDV^T$, where $Z=UD$ are the principal components and $V$ are the loadings of the principal components. The $ith$ principal component $Z_i, i\in (1, N)$ represents the $ith$ mode of variation in the anatomical deformation while the corresponding loadings $V_i$ represent the contribution of each variable to this mode of variation \citep{lee2005assessment}. The $V_i$ calculated by PCA are usually all non-zero values and hence the selected informative vertices are highly related and clustered. The aim of SPCA is to achieve a sparse $V_i$. $V_i$ can be recovered by:

\begin{equation}
\hat{\beta}_{ridge}=arg\hspace{0.2cm} min_{\beta}\|Z_i-Y_{norm}\beta\|^2+\lambda\|\beta\|^2
\label{equ:PCA_regress}
\end{equation}

Here, $\frac{\hat{\beta}_{ridge}}{\|\hat{\beta}_{ridge}\|}=V_i$, $\lambda$ is a manually set positive parameter. With the addition of an $L_1$ constraint, Eq. \ref{equ:PCA_regress} becomes:

\begin{equation}
\hat{\beta}_{ridge}=arg\hspace{0.2cm} min_{\beta}\|Z_i-Y_{norm}\beta\|^2+\lambda\|\beta\|^2+\lambda_1\|\beta\|_1
\label{equ:PCA_constraint}
\end{equation}

where $\|\beta\|_1=\sum_{j=1}^q|\beta_j|$ and $\lambda_1$ is a manually set parameter which controls the sparsity or the number of non-zero values of $\hat{\beta}_{ridge}$. Eq. \ref{equ:PCA_constraint} can be solved with a fixed $\lambda$ and any $\lambda_1$ by least angle regression elastic net (LARS-EN) efficiently \citep{zou2005regularization}.

However, Eq. \ref{equ:PCA_constraint} is still based on PCA due to the inclusion of $Z_i$. To solve this, a two-stage exploratory analysis was formulated with PCA initialization and then optimization with sparse approximations.

With $y_i$ - the $ith$ row of $Y_{norm}$:

\begin{equation}
(\hat{\alpha},\hat{\beta})=arg\hspace{0.2cm} min_{\alpha,\beta}\sum_{i=1}^N\|y_i-\alpha\beta^Ty_i\|^2+\lambda\|\beta\|^2
\label{equ:PCA_PC_1}
\end{equation}

When $\|\alpha\|^2=1$, then $\hat{\beta}\propto V_1$; the detailed proof can be found in \citep{zou2006sparse}. If the first $k$ principal components and the lasso penalty are included, Eq. \ref{equ:PCA_PC_1} becomes

\begin{equation}
(\hat{A},\hat{B})=arg\hspace{0.2cm} min_{A,B}\sum_{i=1}^N\|y_i-AB^Ty_i\|^2+\lambda\sum_{j=1}^k\|\beta_j\|^2+\sum_{j=1}^k \lambda_{1,j}\|\beta\|_1
\end{equation}
\label{equ:PCA_PC_k}

Here, $A_{q\times k}=[\alpha_1,...,\alpha_k]$ are the loadings of the first $k$ principal components of PCA, when $A^TA=I_{k\times k}$. Then $B_{q\times k}=[\beta_1,...,\beta_k]$ are the approximated sparse loadings of $V_{1:k}$. 

The complete SPCA algorithm is listed in Algorithm \ref{tab:SPCA}. The approximated sparse loadings $\hat{V}_j$ is a $q\times 1$ matrix with the loading or contribution of each variable to the $jth$ principal component or mode of variation. The parameter $\lambda_{1,j}$ controls the sparsity or the number of non-zero values in $\hat{V}_j$. As suggested in \citep{lee2005assessment}, the contribution of three coordinates (x, y, z) at $\hat{V}_1$ was added together to represent the vertex contribution. The vertices at all time frames with non-zero contribution were selected as the informative vertices. A plane with the minimum sum of distances to all informative vertices was determined as the optimal scan plane. When calculating the sum, each distance was weighted by the vertex contribution. For multiple scan planes, $\hat{V}_j, j\in(2,N)$ can be used to determine the $jth$ optimal scan plane; however, this is out of the scope of this paper as we are targeting a single scan plane.

In \citep{lee2010dynamic}, the real scan planes were registered to the optimal scan planes. In this paper, as the proposed KPLSR-based 3D shape instantiation is robust to local scan plane deviations, which will be shown in Sec. \ref{sub: Scan Plane Deviation}, the final scan plane is an approximate one that is both accessible and convenient for imaging with parameters near the optimal scan plane. When the deformations or shapes of the anatomy are significantly different between patients and hence there are significantly different optimal scan planes between patients, such as the metastatic liver after oncological surgery, the optimal scan plane needs to be determined on a patient-specific basis. When the deformations or shapes of the anatomy are similar between patients and hence there are similar optimal scan planes between patients, such as the RV, the trend of the optimal scan planes for multiple patients is determined as a general optimal scan plane for the anatomy and will be used directly in subsequent interventional procedures, thus reducing the workload for clinicians. Detailed optimal scan plane determination and approximation in our experiments are given in Sec. \ref{Sec: Data}.

\begin{table}
\centering
\caption{SPCA \citep{zou2006sparse}}
\begin{tabular}{l}
\hline 
\textbf{SPCA} \vspace{0.5mm} \\
\hline 
Initialize $A=V[:,1:k]$:\\
\hspace{0.5cm}the loadings of the first $k$ principal components from PCA\\ 
Initialize $B_{q\times k}=[\beta_1,...,\beta_k]=0$\\
For j=1:k\\
\hspace{0.5cm}If $\|\beta_j^{new}-\beta_j^{old}\|>criterion$, which has not converged\\
\hspace{1cm}Solve the following minimum by LARS-EN: \\
\hspace{1.5cm}$\beta_j=arg\hspace{0.2cm}min_{\beta}(\alpha_j-\beta)^TY_{norm}^TY_{norm}(\alpha_j-\beta)+\lambda\|\beta\|^2+\lambda_{1,j}\|\beta\|_1$\\
\hspace{1cm}Update $B$ with the normalized new $\beta_j$\\
\hspace{1cm}Update $A$ with the normalized new $\alpha_j$:\\
\hspace{1.5cm}$\alpha_j=(1-A[:,1:j-1]A[:,1:j-1]^T)Y_{norm}^TY_{norm}\beta_j$\\
\hspace{0.5cm}End\\
End\\
Approximated sparse loadings $\hat{V}_j=\beta_j, j=1,...,k.$\\
\hline
\end{tabular}
\label{tab:SPCA} 
\end{table}

\subsection{3D Shape Instantiation}
\label{Sec: Prediction}
With the pre-operative 3D SSM and the approximate optimal scan plane obtained, 2D images synchronized with the time frames for pre-operative scanning were obtained at the approximate optimal scan plane. The 2D anatomical contours were segmented and sampled to the same number of 2D vertices and connectivities, resulting in a 2D SSM with vertices $X_{N\times numX\times 2}$, where $N$ is the number of time frames and $numX$ is the number of vertices. By rearranging the (x, y) coordinates of vertices as independent variables, $X_{N\times p}$ was obtained, where $p=numX\times 2$ is the number of variables and typically $p\neq q$. The 3D volumes and 2D images do not need to be registered. KPLSR is then applied to learn the relationship between the 3D SSM which is the response in regression and the 2D SSM which is the predictor in regression. For 3D shape instantiation, the new intra-operative 2D image is obtained at the same scan plane and is sampled into the same number of vertices and connectivities as that for the original 2D SSM with vertices $x'_{numX\times 2}$. $x'_{1\times p}$ is obtained  by rearranging the $(x,y)$ coordinates as independent variables for applying the KPLSR-derived relationship to predict the intra-operative response $y'_{1\times q}$ whose $(x,y,z)$ coordinates are then rearranged back to obtain the intra-operative 3D shape $y'_{numY\times 3}$. 

In the following sections, we introduce PLSR and show its extension to KPLSR.

\subsubsection{PLSR}
PLSR extracts the relationship between two matrices which could have different dimensions. With predictors $X_{N\times p}$ and responses $Y_{N\times q}$, PLSR finds the relationship:

\begin{equation}
\hat{Y}_{{N\times q}}=X_{N\times p}B_{p\times q}
\end{equation}

Here, $\hat{Y}$ is the prediction of $Y$. The latent variables in $X$ are determined by $B_{p\times q}$ to explain the latent variables in $Y$ maximally. $B_{p\times q}$ is later used to predict the intra-operative response $y'_{1\times q}$ from $x'_{1\times p}$. Non-linear iterative partial least squares (NIPALS) is a widely applied PLSR algorithm \citep{rosipal2001kernel}. In this paper, an alternative algorithm - SIMPLS \citep{de1993simpls} - was used for increased time-efficiency.

Without loss of generality, both $X_{N\times p}$ and $Y_{N\times q}$ are centered with the mean of each column as $0$, which are expressed by $X_0$ and $Y_0$ respectively below. The main problem for SIMPLS is to compute the weight factors $r_i$ and $d_i$, where $i\in (1,M)$ and $M$ is a manually set parameter denoting the number of components used. $r_i$ and $d_i$ maximize the covariance of $t_i=X_0r_i$ and $u_i=Y_0d_i$ with the following four conditions:

\vspace{2mm}
1. maximized covariance: $u'_it_i=d'_i(Y'_0X_0)r_i = maximum$, 

2. normalized $r_i$: $r'_ir_i=1$, 

3. normalized $d_i$: $d'_id_i=1$, 

4. orthogonalized $t$: $t'_jt_i=0, i>j$
\vspace{2mm}

To satisfy the fourth condition, $t'_jt_i=t'_jX_0r_i=(t'_jt_j)c'_jr_i=0$, where $c_j=X'_0t_j/(t'_jt_j)$. When $i>1$, any new $r_i$ must be orthogonal to $C_{i-1}=[c_1,c_2...c_{i-1}]$. This orthogonal projector is $I_p-C_{i-1}(C'_{i-1}C_{i-1})^{-1}C'_{i-1}$, where $I_p$ is an identity matrix. The SIMPLS algorithm is listed in Algorithm \ref{tab:SIMPLS}:

\begin{table}
\centering
\caption{SIMPLS}
\begin{tabular}{l}
\hline 
\textbf{SIMPLS} \vspace{0.5mm} \\
\hline
Initialize $S_0=X_0^TY_0$, $X_0,Y_0$ are the centered matrix of $X, Y$ respectively\vspace{0.5mm}\\
for $i=1:M$ (M is a manually set parameter - the number of components used) \\
\hspace{0.5cm} if $i==1$ \\
\hspace{1cm}$r_i=$ first left singular vector of SVD of $S_0$, ($r_i - weights$)\\
\hspace{0.5cm} else \\
\hspace{1cm}$r_i=$ first left singular vector of SVD of $S_0(I_p -C_{i-1}(C'_{i-1}C_{i-1})^{-1}C'_{i-1})$\\ 
\hspace{0.5cm} end \vspace{0.5mm} \\
\hspace{0.5cm}$t_i=X_0r_i$ ($t_i - scores$)\\
\hspace{0.5cm}$c_i=X_0^Tt_i/(t_i^tt_i)$ ($c_i - loadings$)\\
end\vspace{0.5mm}\\
Coefficient: $B_{p\times q}=RT^{-1}Y_0$, where $R=[r_1,r_2...r_M],$ $T=[t_1,t_2...t_M]$ \vspace{0.5mm}\\
\hline
\end{tabular}
\label{tab:SIMPLS} 
\end{table}

\subsubsection{KPLSR}
PLSR is less suitable for regressing non-linear motions. KPLSR was used to compensate for this shortage. A kernel function maps the predictor $X_{N\times q}$ into a new feature space $F$ non-linearly with $\Phi: x_i\in R^q \rightarrow \Phi(x_i)\in F, i\in(1,N)$. $\Phi$ satisfies the \textit{kernel trick}: ${\Phi(x_i)}^T\Phi(x_j)=K(x_i,x_j)$. PLSR is then constructed in the feature space $F$ to achieve a non-linear regression for $X$ \citep{rosipal2001kernel}.

The kernel used in this paper was a Gaussian kernel for its increased accuracy over a polynomial kernel:
\begin{equation}
K_{space}=exp(-K/W)
\end{equation}

Here $K_{(i,j)}=K_{(j,i)}=(x_i-x_j)^2,i,j\in (1,N)$. $W$, the Gaussian width, was adjusted to $Ratio\times maximum(K_{N\times N})$ to facilitate parameter adjustment between different targets and subjects, $Ratio$ is a manually set ratio which we term the Gaussian ratio. Substituting the $X$ in Algorithm \ref{tab:SIMPLS} with $K_{space}$ gives us the algorithm for KPLSR.

\subsection{Data Collection and Validation}
\label{Sec: Data}
The proposed framework was validated on both liver and cardiac RV studies. The experiments included two digital liver phantoms, one dynamic liver phantom, one in vivo porcine liver, eight livers from metastatic patients, 18 cardiac RVs from asymptomatic subjects, and 9 cardiac RVs from HCM patients.

The acquisition of 3D meshes and synchronized 2D contours at different time frames along the dynamic cycle for each data are given in Sec. \ref{Sec: Dynamic Phantom} - \ref{Sec: Cardiac Data}. All data used the same methods to construct the 3D and 2D SSM. With known 3D shapes consisting of 3D vertices and connectivities at different time frames, the mid-state 3D mesh was first projected to meshes at other time frames by non-rigid registration \citep{nonrigidICP}. Then the registered mid-state 3D mesh was mapped onto meshes at other time frames by projecting its vertices along the normal directions. Therefore a 3D SSM with point correspondences was constructed. With known 2D contours consisting of 2D vertices and connectivities at different time frames, the construction of a 2D SSM was in the same way as that for a 3D SSM but with a different registration method \citep{Shapecontext}.

\subsubsection{Digital Livers}
\begin{figure}[thpb]
\centering
\includegraphics[width=1.0\textwidth]{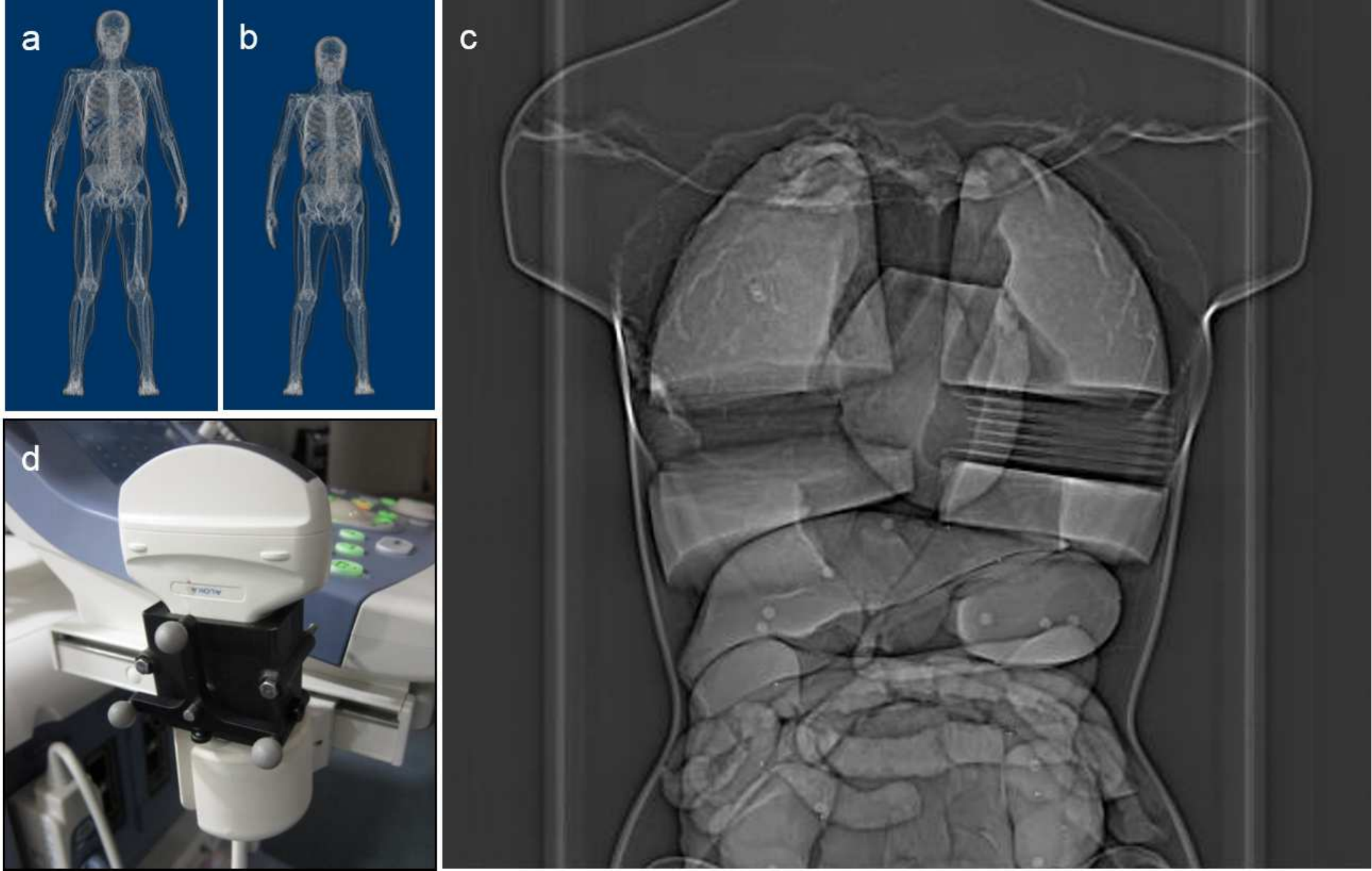}
\caption{The digital livers and phantom experiment setup: (a) the male digital phantom, (b) the female digital phantom, (c) an X-ray image of the Regina phantom, whose lungs have been modified to simulate different respiratory positions, (d) the custom designed tracking frame based on a Polaris tracker mounted on the ultrasound transducer.}
\label{fig:Phantom_Setup}
\end{figure}

XCAT is a digital whole body phantom with detailed, high-resolution and dynamic tissues \citep{segars20104d} as shown in Fig. \ref{fig:Phantom_Setup}a and Fig. \ref{fig:Phantom_Setup}b. In this paper, the isotropic resolution of the volume was set at $0.625mm$. 21 time frames were collected between exhalation and inhalation. 3D meshes of two XCAT livers (one male and one female) were segmented and processed with Analyze (AnalyzeDirect, Inc, Overland Park, KS, USA) and MeshLab \citep{Meshlab}. A 3D SSM was constructed for each digital liver. 

The optimal scan plane for each liver was determined with approximately $200$ informative vertices and was used to slice the meshes in the 3D SSM. The intersection contours were projected onto the slicing plane to simulate 2D contours. A 2D SSM was constructed for each liver.

\subsubsection{Dynamic Liver Phantom Experiment}
\label{Sec: Dynamic Phantom}
A detailed female phantom modeled with silicone organs (the Regina model \citep{lerotic2010multimodal}) was used. The lungs were modified to simulate respiratory motion. In each lung, foam board inserts (each $5mm$ thick) were used, creating seven different liver deformation positions. Each respiratory position was scanned in a Siemens 64 slice SOMATOM™ Sensation CT Scanner with images of $0.77 mm \times 0.77 mm$ in-plane resolution and $1mm$ slice separation. Segmentation and processing were performed with Analyze and MeshLab.

For real-time scanning, ultrasound imaging was used. A 2D imaging transducer used with the ALOKA prosound $\alpha10$ system (Aloka Co. Ltd, Tokyo, Japan) was affixed with an NDI Polaris passive infrared tracker (Northern Digital, Inc, Waterloo, ON, Canada), enabling the recording of the spatial position and orientation of the scan plane. Calibration between systems was established by registering three known landmarks on the liver phantom in both frames of reference. The ultrasound images were captured from the S-video output feed of the scanner. The experiment setup is shown in Fig. \ref{fig:Phantom_Setup}c and  Fig. \ref{fig:Phantom_Setup}d.

Freehand 3D ultrasound systems require calibration and a number of techniques and corresponding phantoms have been developed for this \citep{mercier2005review}. To calibrate the ultrasound images to the coordinate space of a tracking device, a three-point crossed wires phantom was built. The transforms from the coordinate space of the optical tracker to that of the CT imaging space were calculated by PRAXIS \citep{gegenfurtner1992praxis}. This defined a translation and a quaternion for the rotation between the ultrasound image points and the CT imaging space \citep{prager1998rapid}. The mean distance between the registered ultrasound image points and the 3D meshes scanned by CT is less than $10^{-2}mm$.

Due to the constraints caused by the rib cage, the optimal scan plane fitted with 30 informative vertices was selected as the actual scan plane. The silicone phantom was filled with water. For each respiratory position, the optimal scan plane was acquired with the ultrasound probe. An experienced operator scanned the phantom using an in-house guidance system where the silicone liver was registered to a 3D guidance mesh by three manually chosen points. This guidance system provided the actual scan plane in real-time as well as the desired scan plane orientation.

A semi-automatic segmentation based on active contours \citep{geiger1995dynamic} was used to delineate the liver contour from the 2D ultrasound images. It could determine a contour in an ultrasound image automatically when the two end points were selected manually. The contours were transformed to the CT coordinate frame to achieve registered contour coordinates from which a registered 2D SSM was constructed. The registration between 3D volumes and 2D images was only performed for the Regina phantom for later specific comparison and was not performed for all other data. 

\subsubsection{Porcine Liver}
One contrast enhanced 3D CT scan was captured at full exhalation using a GE Innova 4100 interventional X-ray machine capable of fluoro-CT imaging. Due to the respirator design, the porcine liver could not be stopped at different respiratory positions for a 3D CT scan. Instead, fluoroscopic images were obtained in an anteroposterior (AP) direction over time to cover the animal's respiratory motion. As only one 3D volume at full exhalation was scanned with CT, 3D volumes at other respiration positions were simulated by image constrained finite element modeling (FEM) \citep{lerotic2009image} while the collected fluoroscopic images were used as the image constraints. This created multiple liver 3D meshes at different time frames. In this paper, the surface mesh at full exhalation was first turned into a tetrahedral mesh using Gmsh \citep{geuzaine2009gmsh}. Then, the Open Source SOFA framework \citep{allard2007sofa}, chosen for its emphasis on real-time medical simulations, was used for the FEM. The material for the liver was set to be elastic and isotropic, with a Young's modulus of 640 Pa and Poisson's ratio of 0.3 \citep{yeh2002elastic}. A 3D SSM was constructed for the porcine liver.

The meshes in the 3D porcine liver SSM were sliced by the optimal scan plane determined with approximately $200$ informative vertices. The sliced contours were projected onto the slicing plane with 2D coordinates to simulate 2D contours. A 2D SSM was constructed for the porcine liver as well.

\subsubsection{Metastatic Livers}
Clinical data from eight patients (6 male, 2 female, mean age 63) with metastatic liver tumors was collected. 4D volumes were scanned using a 1.5T MR scanner (Intera, Philips, Amsterdam, Netherland) using a T1 weighted free-breathing sequence (TR = $7.83 ms$, TE = $2.24 ms$, $3.5mm\times3.5mm$ in-plane resolution, $4.5mm$ slice thickness). Each volume consisted of $45$ slices and was acquired in approximately $1.2s$. $60$ time frames were collected to cover the liver motion during respiration. Due to motion artifacts caused by respiration, we could only confidently segment the livers at full inhalation and full exhalation. As before, the SOFA framework was used to generate the meshes at different respiratory positions but with the 3D volumes at full inhalation and full exhalation as the constraints. These meshes were used to construct a 3D SSM for each patient.

\begin{figure}[thpb]
\centering
\includegraphics[width=1.0\textwidth]{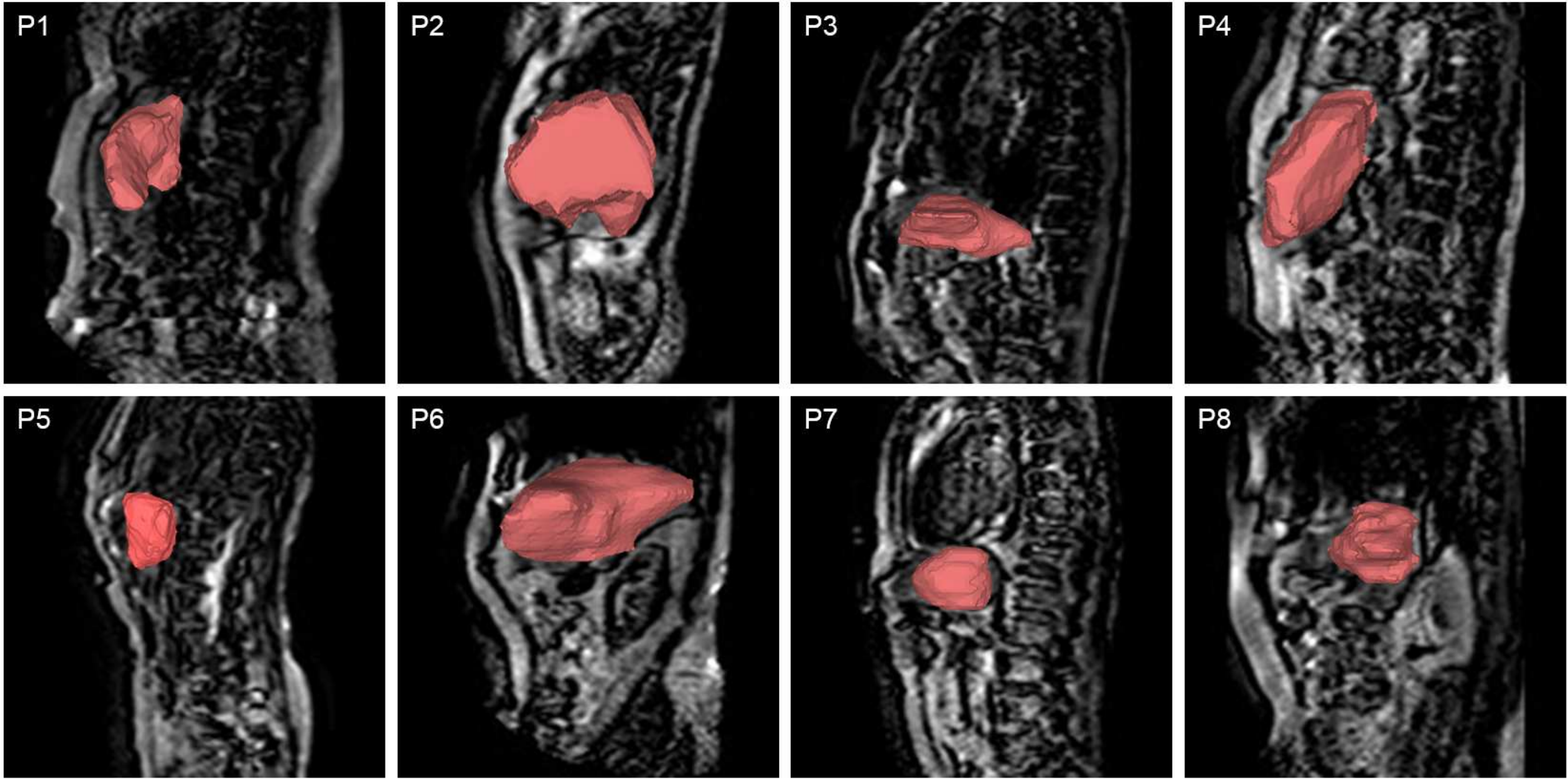}
\caption{The eight metastatic livers shown on their corresponding MR images.}
\label{fig:Liver_Patient_Illustration}
\end{figure}

The eight metastatic patients have significantly different liver shapes and deformations, as shown in Fig. \ref{fig:Liver_Patient_Illustration} and the optimal scan planes for each patient were very different. For this reason, for the metastatic liver, the optimal scan plane was determined patient-specifically with approximately $50$ informative vertices for each patient and this was used to slice the meshes in the 3D SSM. The sliced contours were projected onto the slicing plane with 2D coordinates to simulate 2D contours. A 2D SSM was constructed for each patient.

\subsubsection{Cardiac Data}
\label{Sec: Cardiac Data}
27 subjects (18 asymptomatic subjects (Subjects 1-18) and 9 patients with HCM (Subjects 19-27)) were scanned with a 1.5T MR scanner (Sonata, Siemens, Erlangen, Germany). HCM was selected as it is one of the diseases that influence both the shape and deformation of the heart significantly. Short-axis cine sequences from the atrioventricular ring to the apex were scanned with a $10mm$ slice gap and a $1.5-2mm$ pixel spacing. $19-25$ time frames were collected. To recognize the slice location of the atrioventricular ring and the apex, the $10mm$ slice gap was interpolated to $1mm$ in Analyze. 3D RV meshes were segmented and built with Analyze and MeshLab. A 3D SSM was constructed for each patient.

\begin{figure}[thpb]
\centering
\includegraphics[width=1.0\textwidth]{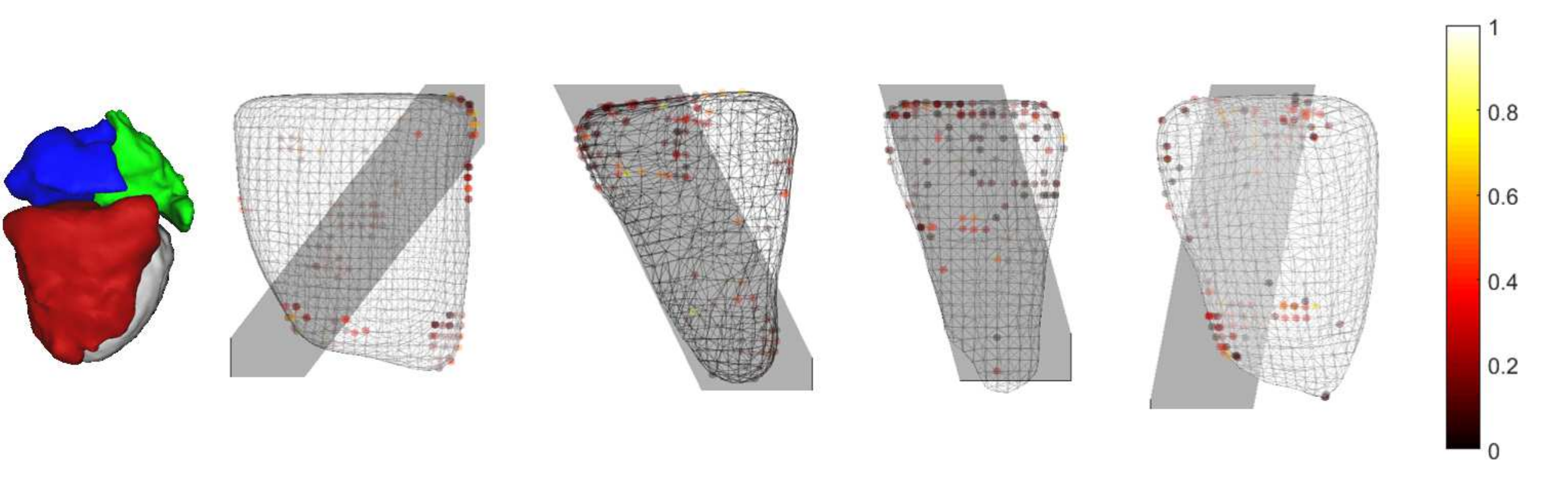}
\caption{Four RVs, with optimal scan plane determination using the $150$ most informative vertices: the vertices are colored by their normalized importance determined by SPCA and the grey plane is the optimal scan plane, with the overall view direction shown on the left hand side. The red/blue/green/grey chambers are the right ventricle/right atrium/left atrium/left ventricle, respectively.}
\label{fig:SPCA}
\end{figure}

Even though HCM influences both the shape and deformation of the RV, the optimal scan planes for the 27 subjects, which were determined with approximate $150$ informative vertices each, were mostly found to be along the long axis of the heart. Four examples are shown in Fig. \ref{fig:SPCA}. Even though the optimal scan planes in Fig. \ref{fig:SPCA} are not exactly the same, they still share the same trend $\textendash$ lying along the long axis of the heart. This similarity of the optimal scan planes between patients is mainly due to the similarity in deformation and shape of the RVs between patients. As later KPLSR-based 3D shape instantiation is robust to optimal scan plane deviations, we made an adjustment to the optimal scan plane to ensure the accessibility of the scan plane and the visibility of the RV considering the following three issues: 1) The long-axis is accessible for 2D MR, 2) The horizontal (four-chamber) long-axis has a clear view of the RV without overlap with other chambers, and 3) Clinicians are familiar with this plane as it features the apex and the atrioventricular ring. For these reasons, the horizontal (four-chamber) long-axis plane was selected as the actual scan plane for all RVs.  

2D MR images at the horizontal (four-chamber) long-axis plane with the synchronized time frames as that for the pre-operative 4D volume scanning were obtained for all 27 subjects. Analyze was used to segment the RV contours and a 2D SSM was constructed for each subject.

\subsubsection{Validation}
Leave-one-out cross validation was applied for all time frames for all data. The $ith, i\in (1,N)$ time frame in the 2D SSM was left out as a new predictor while the $ith$ time frame in the 3D SSM was left out as the ground truth. All other time frames were used in the learning. The error was calculated as the Euclidean vertex-to-vertex distance between the 3D prediction and the ground truth. The shape variation was calculated as the mean vertex-to-vertex distance between the $(i-1)th$ and the $(i+1)th$ time frame in the 3D SSM.

It was shown that SPCA was able to better select the real and unrelated informative variables than PCA on a synthetic example \citep{zou2006sparse}. For the synthetic example, the contribution of a variable and the relations between variables were known. However, for practical data, both this contribution and the relations were unknown; a comparison of the distribution of the informative vertices selected by PCA and SPCA is given in this paper. In practice, adjusting the optimal scan plane is usually necessary for better scan plane accessibility and target visibility. To illustrate that this adjustment will not incur major errors, multiple deviated optimal scan planes were used to instantiate the 3D shape.

Both PLSR and KPLSR regress the relationships between two matrices rather than two coordinate frames. Lee et al. applied PLSR with registration of pre-operative 3D SSM and synchronized 2D SSM \citep{lee2010dynamic}. In this paper, this explicit registration is not required. To demonstrate this, both the registered and non-registered 2D SSM of the dynamic phantom liver were used as the predictor for dynamic shape instantiation. The stability of an algorithm with respect to its parameters is important for judging its performance. PLSR has one parameter, the number of components used, while KPLSR has two parameters, the number of components used and the Gaussian ratio. To evaluate the stability of PLSR and KPLSR to the number of components used, the validation was applied on two HCM RVs with the number of components used set from $1-18$. In practice, the time frames at or near the boundaries, i.e. at maximal inhalation and exhalation or at diastole and systole, are the most difficult time frames to recover, as the learning is more weak for these time frames. We term these time frames \textit{boundary time frames}. In this paper, the performance of PLSR and KPLSR at boundary time frames were validated on two cardiac RVs (one asymptomatic RV and one HCM RV). The liver data was collected along half of the dynamic cycle - the first and last few time frames are the inhalation and exhalation respectively, i.e. the boundary time frames. The cardiac data was collected along the entire dynamic cycle, the first and last few time frames are at diastole while the middle few time frames are at systole, i.e. the boundary time frames. 

Finally, the accuracy of the proposed dynamic shape instantiation was tested on two digital livers, one in vivo porcine liver, eight metastatic liver patients, and 27 RVs of asymptomatic subjects and HCM patients.

\section{Results}
The results from our experiments are shown in the following sections. The comparison between PCA and SPCA on selecting informative vertices is demonstrated in Sec. \ref{sub: PCA and SPCA}. The robustness of the KPLSR-based 3D shape instantiation to scan plane deviations is shown in Sec. \ref{sub: Scan Plane Deviation}. The validation on releasing the registration between pre-operative 3D SSM and synchronized 2D SSM is illustrated in Sec. \ref{sub: Registration}. The stability of PLSR and KPLSR to the number of components used is compared in Sec. \ref{sub: Stability to the Number of Components Used}. Boundary time frames are tested in Sec. \ref{sub: Stability to Boundary Conditions}. Finally, the accuracy of the proposed dynamic shape instantiation is validated on the liver and the heart, which is shown in Sec. \ref{sub:Dynamic Shape Instantiation}.

\subsection{Comparison between PCA and SPCA}
\label{sub: PCA and SPCA}
\begin{figure}[thpb]
\centering
\includegraphics[width=1.0\textwidth]{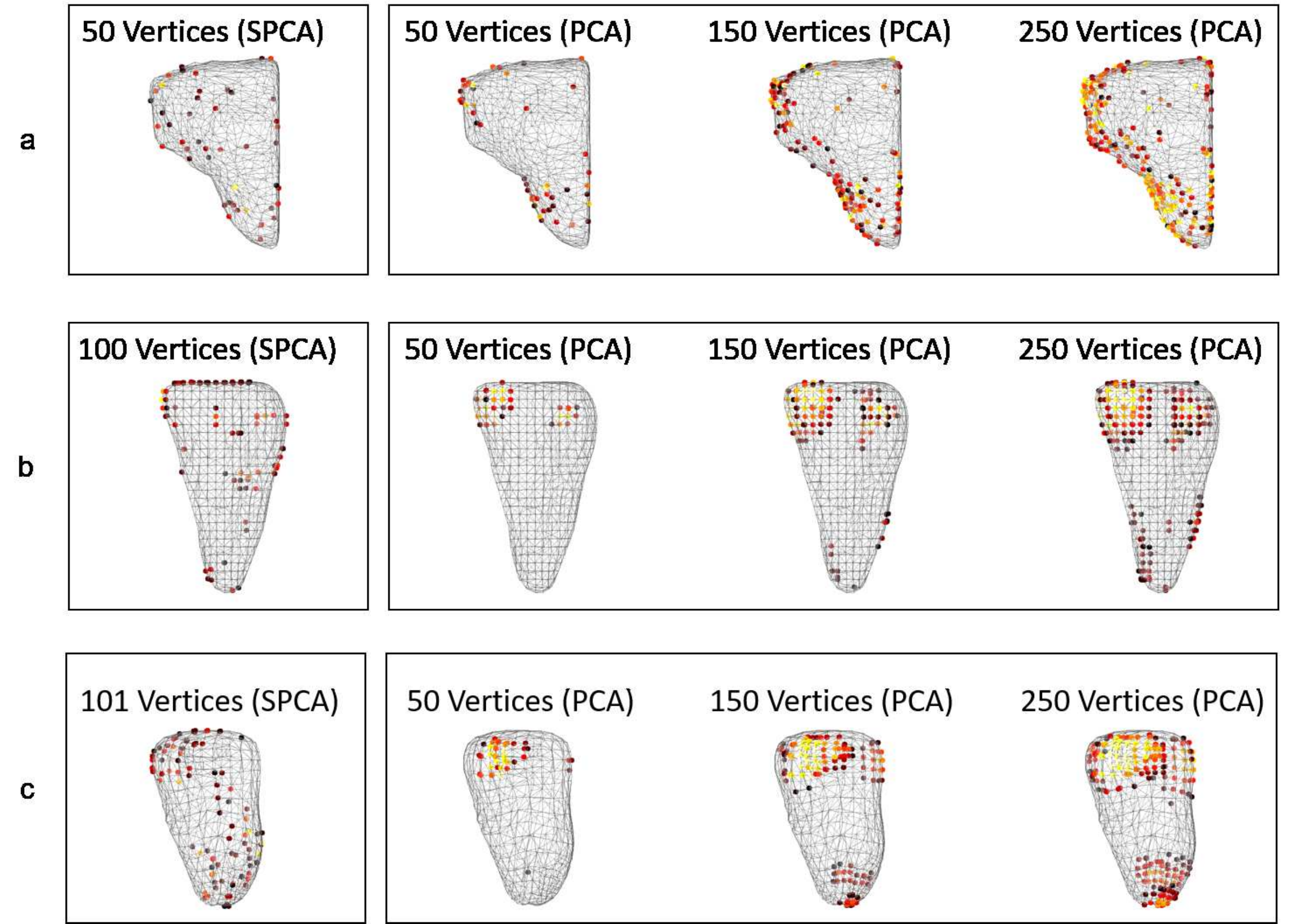}
\caption{One liver and two RV examples showing the most informative vertices selected by SPCA and PCA: (a) a metastatic liver with 50 informative vertices determined by SPCA while 50, 150, 250 informative vertices determined by PCA, (b) an asymptomatic RV with 100 informative vertices determined by SPCA while 50, 150, 250 informative vertices determined by PCA, (c) a HCM RV with 101 informative vertices determined by SPCA while 50, 150, 250 informative vertices determined by PCA. The view directions for RVs and vertex coloring are in the same way as that in Fig. \ref{fig:SPCA}}
\label{fig:PCA_SPCA}
\end{figure}

For most subjects, including the metastatic livers and cardiac RVs, it was found that the informative vertices selected by PCA were more clustered than the informative vertices selected by SPCA. Three examples are shown in Fig. \ref{fig:PCA_SPCA}. Clustered informative vertices were selected by PCA due to their related motion with the informative vertices considered to be in the same area. SPCA can remove this inter-relation and only select the true and sparse informative vertices.

\subsection{Robustness to Scan Plane Deviations}
\label{sub: Scan Plane Deviation}
\begin{figure}[thpb]
\centering
\includegraphics[width=1.0 \textwidth]{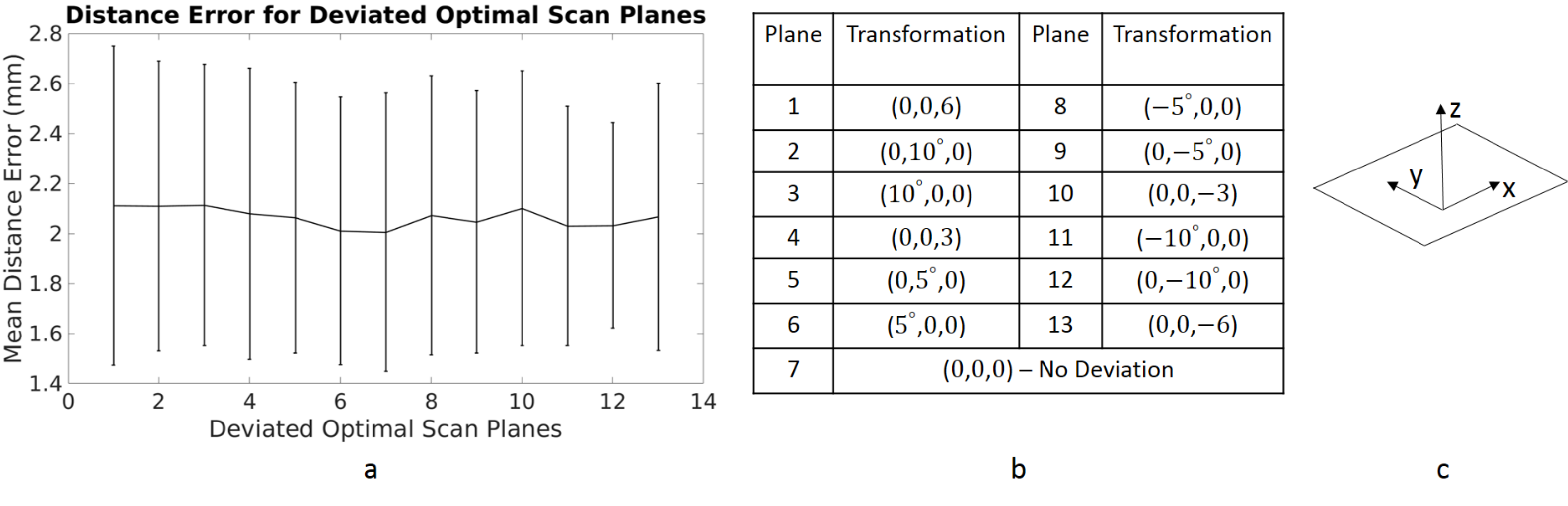}
\caption{Testing the robustness of the proposed KPLSR-based 3D shape instantiation to scan plane deviations: (a) the mean distance error of the 3D shape instantiation with deviated optimal scan planes, with standard deviation calculated across 20 time frames, (b) the deviations of the scan planes. Even though a plane could have six transformations, three of them (rotation along the z axis, translation along the x axis and translation along the y axis do not influence the slicing results. The other three transformations were explored. For example, $(0,0,6)$ means rotating $0^\circ$ along the x axis, rotating $0^\circ$ along the y axis, and translating $6mm$ along the z axis, (c) illustration of the $x, y, z$ axes of a plane.}
\label{fig: DeviatedPlanes}
\end{figure}

To demonstrate the robustness of the proposed KPLSR-based 3D shape instantiation to scan plane deviations, example RV results from Subject 3 are illustrated below. 13 scan planes with some deviations from the optimal scan plane were used to slice the pre-operative 3D SSM for 3D shape instantiation. The distance error and the deviation for each scan plane is shown in Fig. \ref{fig: DeviatedPlanes}a and Fig. \ref{fig: DeviatedPlanes}b respectively. We can see that the achieved accuracy was scarcely influenced by local scan plane deviations, demonstrating the robustness of the proposed KPLSR-based 3D shape instantiation to scan plane deviations. This is important for practically implementing the proposed framework, as due to practical constraints (access window, or other local, physical constraints), it may be necessary to deviate slightly from the theoretical optimal scan plane. Such deviation should not induce large changes in instantiation errors.

\subsection{Validation of Registration-Free Instantiation}
\label{sub: Registration}
\begin{figure}[thpb]
\centering
\includegraphics[width=1.0\textwidth]{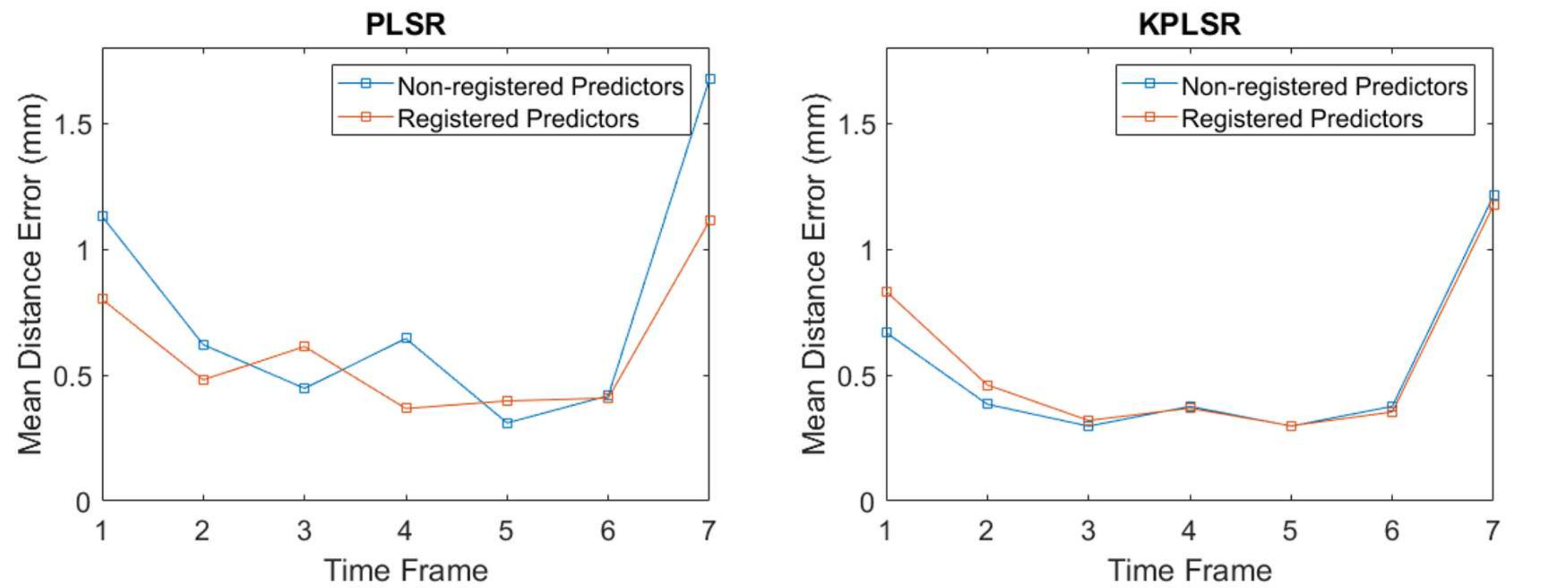}
\caption{The instantiation accuracy for the liver phantom experiment: (left) the mean distance errors of PLSR with registered and non-registered predictors, (right) the mean distance errors of KPLSR with registered and non-registered predictors.}
\label{fig:With_Without_Registration}
\end{figure}

The instantiation accuracy across all time frames with registered and non-registered predictors which were collected in the liver phantom experiment is shown in Fig. \ref{fig:With_Without_Registration}. It can be seen that PLSR is influenced by the registration while KPLSR shows little influence, demonstrating that explicit registration is not required in the proposed method.

\subsection{Stability to the Number of Components Used}
\label{sub: Stability to the Number of Components Used}
\begin{figure}[thpb]
\centering
\includegraphics[width=1.0\textwidth]{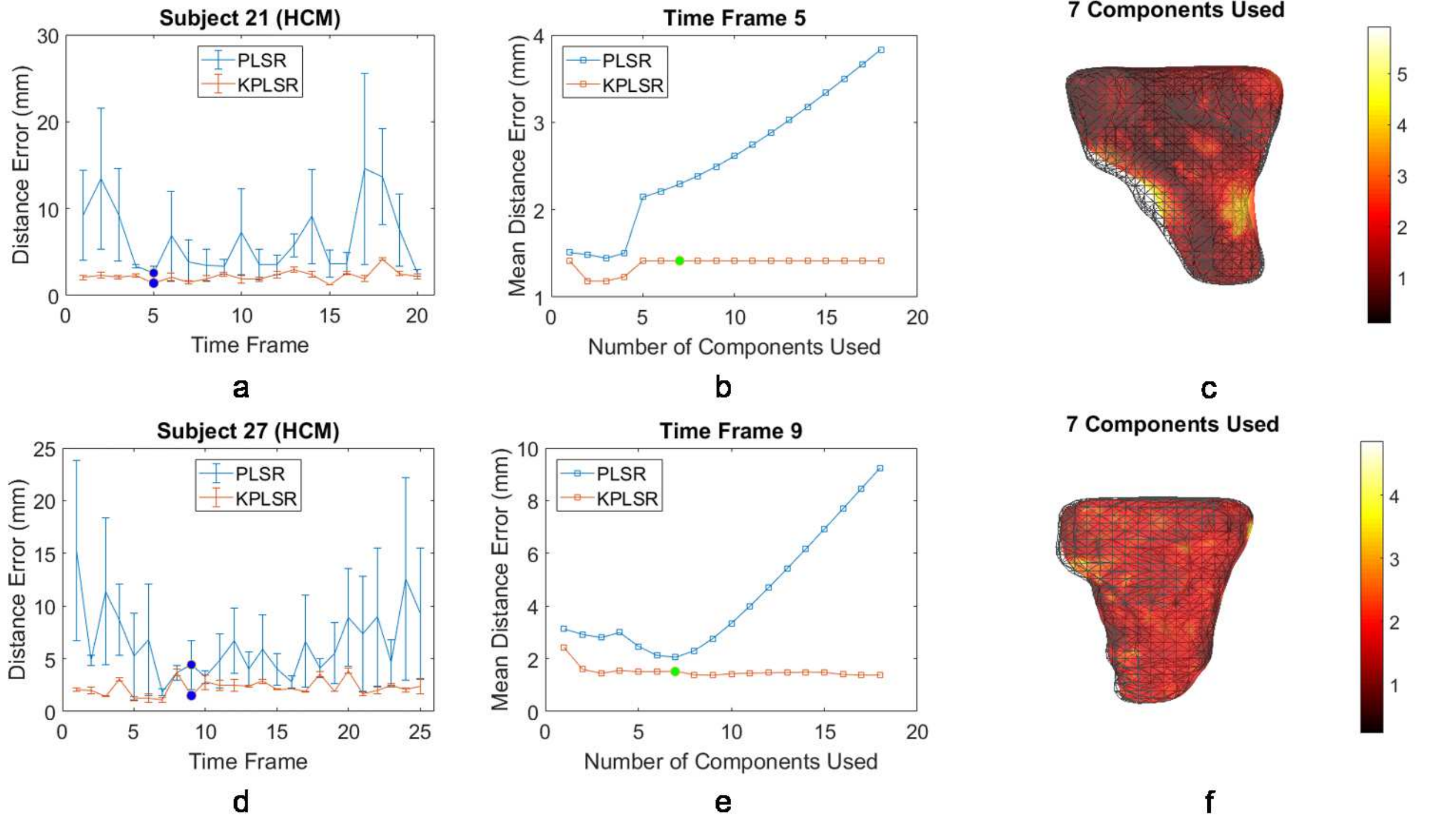}
\caption{Testing the influence of the number of components used on PLSR and KPLSR: (a) the mean $\pm$ std errors for Subject 21, with the standard deviation calculated across $1-18$ components used, (b) mean distance errors with numbers of components used varying from $1-18$ for time frame 5 of Subject 21 (labeled with blue dots in a), (c) a shape instantiation example colored by the distance errors with $7$ components used for time frame 5 of Subject 21 (labeled with green dot in b), with the same view direction in Fig.\ref{fig:SPCA}, d,e,f are the same as a,b,c but for Patient $27$, time frame 9, 7 components used respectively.}
\label{fig:Com}
\end{figure}

Instantiation for two HCM patients (Subject 21 and Subject 27) was calculated along all time frames with a varying number of components used ($1-18$), as shown in Fig. \ref{fig:Com}a and Fig. \ref{fig:Com}d. It can be seen that the accuracy of KPLSR is less sensitive to this parameter - the number of components used - than that of PLSR, as the standard deviations of KPLSR are less than that of PLSR at most time frames. Two time frames (time frame $5$ of Patient 21, time frame $9$ of Patient 27) are shown with the mean distance errors at different numbers of components used in Fig. \ref{fig:Com}b and Fig. \ref{fig:Com}e. Two instantiation examples colored by the distance errors are shown in Fig. \ref{fig:Com}c and Fig. \ref{fig:Com}f. The error is distributed evenly over the mesh.

\subsection{Performance at Boundary Time Frames}
\label{sub: Stability to Boundary Conditions}
\begin{figure}[thpb]
\centering
\includegraphics[width=1.0\textwidth]{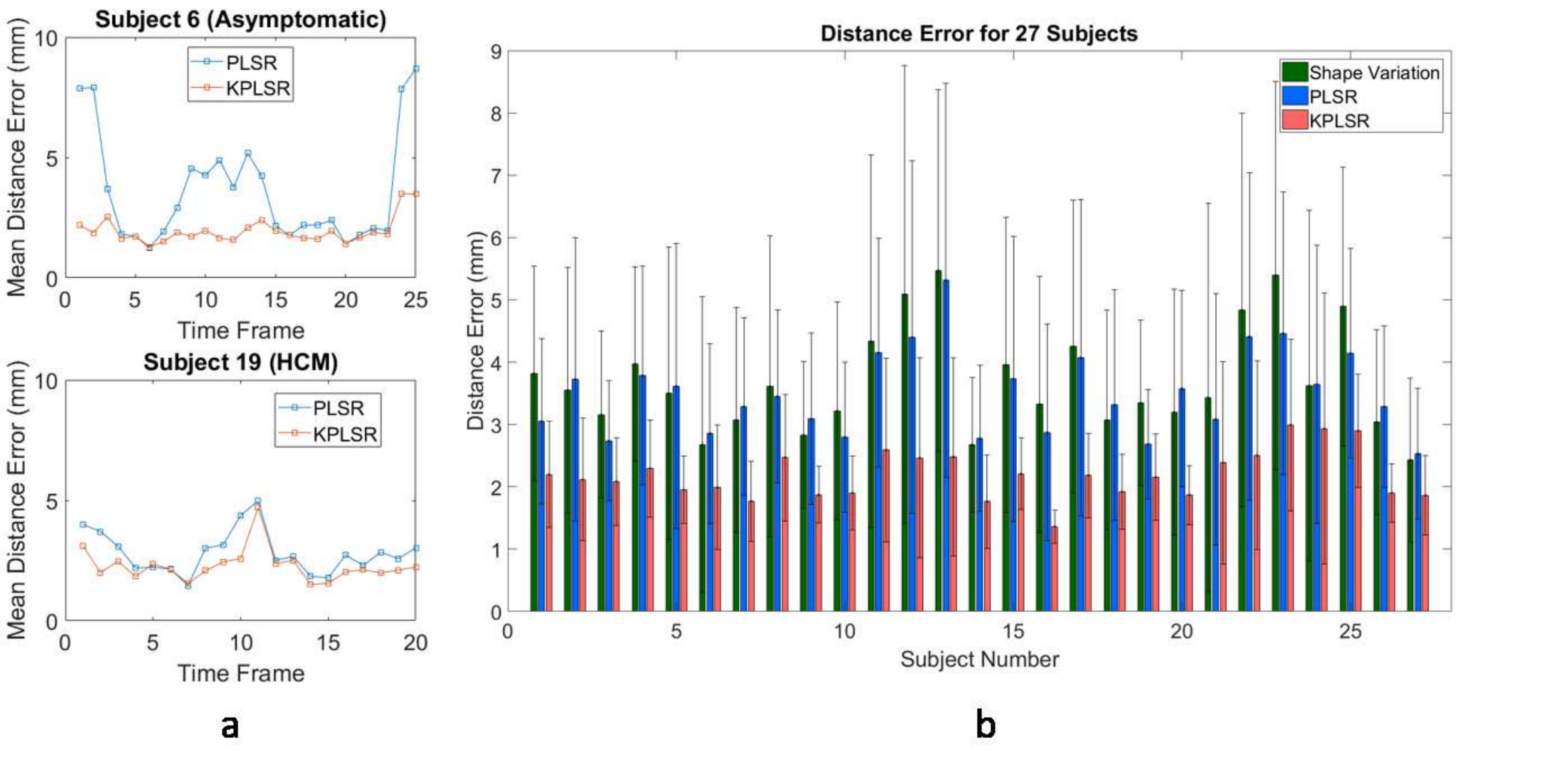}
\caption{Results at the boundary time frames and for the RV experiments: (a) performance test for boundary time frames, (b) the instantiation errors for 27 subjects (Subjects 1-18 = asymptomatic subjects; Subjects 19-27 = HCM).}
\label{fig:partial}
\end{figure}

The mean distance errors for shape instantiation along all time frames are shown for two selected subjects (Subject $6$ (asymptomatic) and Subject $19$ (HCM)) in Fig. \ref{fig:partial}a. The PLSR errors show large peaks near systole (time frame 10 for Subject 6, time frame 11 for Subject 19) and diastole (time frame $1$ and $25$ for Subject 6, time frame $1$ and $20$ for Subject 19) while KPLSR errors show smaller increasing errors at these boundary time frames. It can be concluded that KPLSR has better performance at boundary time frames than PLSR.

\subsection{Accuracy of Dynamic Shape Instantiation}
\label{sub:Dynamic Shape Instantiation}
\begin{figure}[thpb]
\centering
\includegraphics[width=1.0\textwidth]{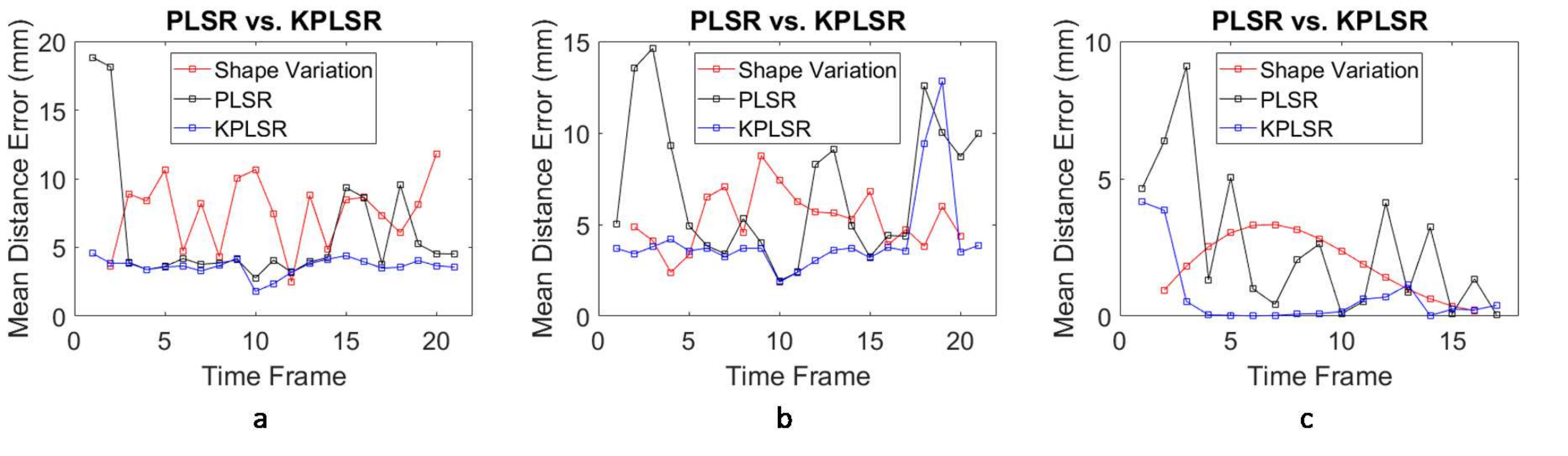}
\caption{The mean distance errors and the shape variation of the two digital livers and the porcine liver: (a) the mean distance errors and the shape variation for the female digital liver, (b) the errors and shape variation for the male digital liver, (c) the errors and shape variation for the porcine liver.}
\label{fig:Liver_Digital_Animal}
\end{figure}

Mean distance errors of PLSR and KPLSR and the shape variation of two digital phantom livers and one porcine liver are shown along all time frames in Fig. \ref{fig:Liver_Digital_Animal}. For the two digital livers, KPLSR achieved much lower errors at the time frames where PLSR showed high peaks. For the porcine data, the accuracy of KPLSR is higher than that of PLSR at most time frames. For both digital phantom and porcine liver studies, the mean distance error of KPLSR is much lower than the shape variation at most time frames. The peaks for KPLSR (time frames 18-19 in Fig. \ref{fig:Liver_Digital_Animal}b and time frames 1-2 in Fig. \ref{fig:Liver_Digital_Animal}c) were caused by boundary time frames. The higher accuracy of KPLSR for the two digital livers is not as obvious as that for the porcine liver due to the design and linear deformation of the digital phantom. 

\begin{figure}[thpb]
\centering
\includegraphics[width=1.0\textwidth]{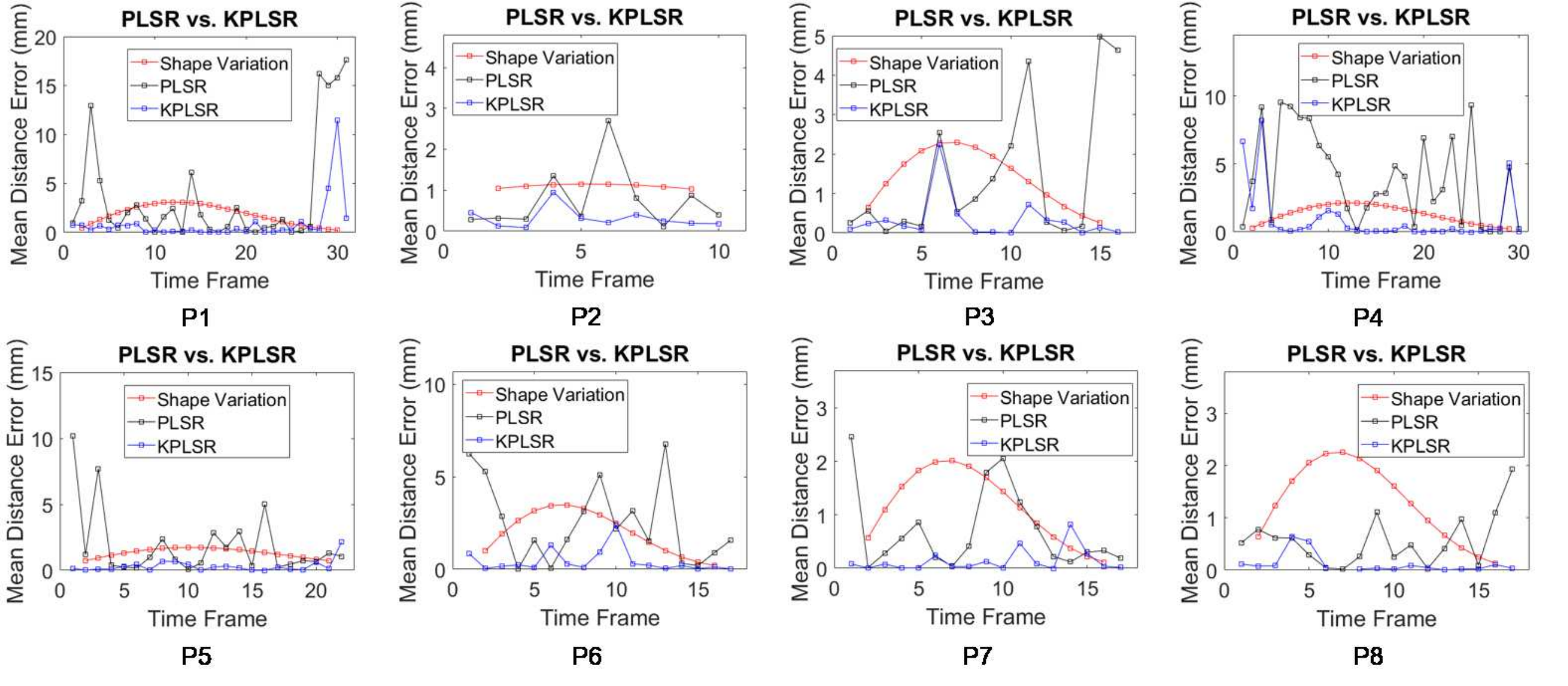}
\caption{The mean distance errors and the shape variation for the eight metastatic livers.}
\label{fig:Liver_Patient}
\end{figure}

Eight patients with metastatic tumors were used for instantiation validation with the mean distance errors of PLSR and KPLSR and the shape variation shown along all time frames in Fig. \ref{fig:Liver_Patient}. For most of the time frames and patients, KPLSR achieved much more accurate instantiation results than those of PLSR. The mean distance errors of KPLSR were also much lower than the shape variation. The higher errors of KPLSR (time frames 29-30 for P1, time frames 1-3 and 29 for P4, time frame 22 for P5, time frame 14 for P7) were caused by boundary time frames.

Shape instantiation of 27 RVs was validated with the mean distance errors of PLSR and KPLSR and the shape variation shown in Fig. \ref{fig:partial}b; the standard deviation in the graphs was from the error variation along different time frames. Overall, KPLSR achieves both lower mean and standard deviation errors in the instantiation than PLSR for all subjects. The error achieved by KPLSR was also much lower than the shape variation for all subjects. The similar results between patients also demonstrate the availability of using one approximate optimal scan plane - the horizontal (four-chamber) long-axis plane for all RVs in this paper.

For optimal scan plane determination, the number of informative vertices was determined as $5\%-10\%$ of the total number of vertices in each test mesh, the parameter $\lambda$ was fixed at 0.0001, $k$ was set at $1$ as we are targeting a single scan plane, and the parameter $\lambda_1$ was set as the number of informative vertices. For 3D shape instantiation, with the exception of the test for stability to the number of components used, all tests were validated with the number of components used for PLSR optimized between $1-8$ while that for KPLSR was empirically set between $1-18$. Overall, KPLSR achieved better accuracy at a higher number of components used than PLSR. The Gaussian ratio parameter of KPLSR was selected empirically. 

Experiments were performed in MATLAB on an Intel(R) Core(TM) i7-4790 CPU @3.60Hz computer. The training took approximately $1s$ for one component deflation; the number of component deflations is the number of components used. The prediction or shape instantiation took approximately $1ms$.

\section{Discussion}
In this paper, SPCA was applied instead of PCA to determine the informative vertices to find the optimal scan plane. We expect that nearby points on the surface of organs will tend to move dependently in a similar fashion. This is because the movement of one cell will cause the movement of its nearby cells due to the connectivity of tissues. the sparse informative vertices determined by SPCA and the clustered informative vertices determined by PCA could illustrate the ability of SPCA to derive principal components from unrelated original variables and hence select the true, unrelated informative vertices. However, from our experiments, the overall trend of the informative vertices selected by PCA was shown to be similar to the trend determined by SPCA. It is more reasonable to conclude that SPCA facilitates the determination of the optimal scan plane more clearly and quickly than PCA rather than more accurately in this case. Setting a higher number of informative vertices when applying PCA could also achieve a good scan plane. 

In practical applications, the calculated optimal scan plane is not always accessible. The robustness of the proposed KPLSR-based 3D shape instantiation to local scan plan deviations ensures the adjustment of the scan plane for better accessibility and visibility in practical clinical scenarios. The optimal scan plane for the RV, which will be used directly for future patients, was determined by analyzing the pattern of the optimal scan planes for 27 RVs. This method of determining the optimal scan plane could be adopted for other anatomies which share similar deformation and shape across patients. For anatomy such as the metastatic liver which has significantly different deformation and shape between patients, the optimal scan plane has to be determined for each patient individually.

The registration between the pre-operative 3D SSM and synchronized 2D SSM is no longer required in this paper. The validation on a liver phantom experiment with both registered and non-registered predictors showed that the accuracy of KPLSR was not influenced by this. The removal of explicit registration will decrease the workload for clinicians significantly when applying the proposed method in practice. It was proved that KPLSR had much higher stability to the number of components used than PLSR. This is important during practical applications in case of the use of a suboptimal setting of this parameter. KPLSR also had better processing at boundary time frames than PLSR though the errors of KPLSR at boundary time frames are still higher than at other normal time frames. This boundary limitation corresponded to more time frames for the liver data than the cardiac data, as the SOFA framework generated meshes at the first few and last few time frames with very small shape variations which were usually less than $0.3mm$. In practical applications, always including the time frames at maximum inhalation and exhalation or at systole and diastole in the training data is highly recommended.

As pre-operative 4D volumes are not typically acquired for livers, FEM was applied to simulate the meshes between the inspiration and expiration. FEM or any other methods which could simulate the physical organ motion can thus be used to generate pre-operative 4D volumes when transferring the proposed framework onto other target anatomies whose dynamic motion is difficult to gate.

In general, three kinds of data are needed to apply the proposed 3D shape instantiation: the 3D SSM for learning, the 2D SSM for learning, and the 2D intra-operative images for prediction. Synchronization is needed between the learning 3D SSM and the learning 2D SSM while registration is needed between the learning 2D SSM and the 2D intra-operative images for prediction. The 4D volume used for constructing the learning 3D SSM was scanned pre-operatively while the 2D images used for constructing the learning 2D SSM could be scanned pre-operatively or intra-operatively, as the learning only takes a few seconds. In practical applications, for organs whose motion could be gated easily, i.e. the RV, the synchronization between the learning 3D SSM and the learning 2D SSM could be achieved through dynamic motion gating, i.e. electrocardiogram (ECG) gating or respiratory gating. The registration between the learning 2D SSM and the intra-operative 2D images for prediction could be achieved by setting the scan machine at the same scan position. For organs whose motion is difficult to gate, i.e. the liver, FEM or other available methods which could simulate the 3D volumes at different time positions could be used to collect the learning 3D SSM and to slice for the learning 2D SSM. The registration between the learning 2D SSM and the intra-operative 2D images for prediction could be achieved by setting the scan machine to the same scan position as that used to slice the learning 2D SSM. 

Two digital livers, one porcine liver, and eight metastatic livers were used to illustrate the applicability of the proposed method on livers. As well, 27 RVs were used in our validation with real 2D MR images as the predictors, which demonstrates the potential value of the proposed method in practical operations. Even with only a single scan plane, a mean distance error of about $2.19mm$ was achieved for the RV. This error was comparable to the mean accuracy in \citep{gao2012registration} and \citep{huang2009dynamic} which were approximately $2.83mm$ and $3.55mm$ for patients and animals, respectively. The computation time for prediction ($1ms$) demonstrates the real-time ability of the proposed method.

\section{Conclusions}
In conclusion, a real-time and registration-free framework for dynamic shape instantiation which is generalizable to multiple anatomies is proposed in this paper. SPCA is applied to select the unrelated and real informative vertices from a pre-operative 3D SSM, which facilitates a more clear and quick determination for the optimal scan plane. KPLSR is used to improve the accuracy and robustness of the instantiation. For anatomies like the RV, the optimal scan plane only needs to be determined once and then can be used in subsequent interventions. The detailed experiments performed for the removal of explicit registration, the stability to the number of components used, and the performance at boundary time frames covers the issues which may occur during practical applications. FEM extends the application of the framework to anatomies like the liver, whose motion is difficult to gate. The patient-specific learning removes the restrictions on the applicable anatomy. This paper sets the basis for applying the proposed framework to other interventional procedures involving dynamic anatomies.

\section{Acknowledgements}
We would like to thank Dr Karim Lekadir and Dr Robert Merrifield for supplying the cardiac data, Dr Maria Hawkins and Dr Diana Tait for the liver patient data, and Dr Mirna Lerotic for her assistance with the finite element simulations. This research was partly supported by the Engineering and Physical Sciences Research Council UK (EP/L020688/1).

\section*{References}
\bibliography{mybibfile}

\end{document}